\documentclass[runningheads,a4paper]{llncs}

\usepackage{amssymb,amsmath,mathrsfs,stmaryrd}
\usepackage[utf8]{inputenc}
\usepackage{multirow}
\usepackage{tikz}
\usepackage{bussproofs}

  {\gdef\scalefactor{#1}\begin{center}\centering\proofSkipAmount \leavevmode}%
  {\scalebox{\scalefactor}{\DisplayProof}\proofSkipAmount \end{center} }

\newcommand{\bicond}{\,\longleftrightarrow\,}
\newcommand{\imp}{\,\longrightarrow\,}
\newcommand{\N}{\mathbb{N}}
\newcommand{\B}{\mathbb{B}}

\begin{document}

\mainmatter  
\title{Automating change of representation for proofs in discrete mathematics.\thanks{The final publication is available at http://link.springer.com.}}

\author{Daniel Raggi\inst{1}\thanks{This work has been supported by a scholarship from the Mexican Council of Science and Technology (CONACYT).} \and Alan Bundy\inst{1} \and Gudmund Grov\inst{2} \and Alison Pease\inst{3}
}

\institute{School of Informatics, University of Edinburgh \and School of Mathematical \& Computer Sciences, Heriot-Watt University \and School of Computing, University of Dundee}

\maketitle

\begin{abstract}
Representation determines how we can reason about a specific problem. Sometimes one representation helps us find a proof more easily than others. Most current automated reasoning tools focus on reasoning within one representation. There is, therefore, a need for the development of better tools to mechanise and automate formal and logically sound changes of representation. 

In this paper we look at examples of representational transformations in discrete mathematics, and show how we have used Isabelle's Transfer tool to automate the use of these transformations in proofs. We give a brief overview of a general theory of transformations that we consider appropriate for thinking about the matter, and we explain how it relates to the Transfer package. We show our progress towards developing a general tactic that incorporates the automatic search for representation within the proving process.
\end{abstract}

\vspace*{-.15cm}
\section{Introduction}
\vspace*{-.15cm}
Many mathematical proofs involve a change of representation from a domain in which it is difficult to reason about the entities in question to one in which some aspects essential to the proof become evident and the proof falls out naturally.

Many times the transformation makes it explicitly into the written proof, but sometimes it remains hidden as part of the esoteric process of coming up with the proof in the mathematician's mind. For a formal, mechanical proof, this can be problematic, not only because we need to account for the logical validity of the transformation, but because if we want a computational system to find a proof like a mathematician would, we need to be able to incorporate something like the esoteric transformations going on inside the mathematician's mind into the mechanical search.

The importance of representational changes in mathematics is evidenced in historically notable works like Kurt Gödel's incompleteness theorems, where the proof involves matching (or encoding) meta-theoretical concepts like `sentence' and `proof' as natural numbers, or more recently Andrew Wiles' proof of Fermat's Last Theorem, which involves matching the Galois representations of elliptic curves with modular forms. This phenomenon is also seen in refinement based formal methods (e.g. VDM and B): one starts with a highly abstract representation that is easy to reason with, and then it is step-wise refined to a very concrete representation that can be implemented as a computer program. All of these transformations are justified by a general notion of morphism.

In this paper we give an overview of a general mathematical framework suitable for reasoning about representational changes in type-theoretic higher-order logics (these are transformations/morphisms between structures that land us in different theories). We see that the operation of Isabelle's \textit{transfer} methods \cite{huffman2013lifting} fit into this notion of transformation. It is a way of mechanising inference between two domains, if the system is provided with a transformation by the user. We present a set of transformations we have identified as essential for reasoning in discrete mathematics, and show how we have used the \textit{transfer} tool to implement mechanical proofs in Isabelle that use these transformations. We show our work towards automating the search for representation as a tactic for use within proofs in discrete mathematics in Isabelle.

\vspace*{-.15cm}
\section{Background}
\vspace*{-.2cm}
Isabelle/HOL is a theorem proving framework based on a simple type-theoretical higher-order logic \cite{isabelle2013}. It is one of the most widely used proof assistants for the mechanisation of proofs. Apart from ensuring the correctness of proofs written in its formal language, Isabelle has powerful automatic tactics like \texttt{simp} and \texttt{auto}, and through time it has been enriched with some internally-verified theorem provers like \texttt{metis} \cite{Hurd2005} and \texttt{smt} \cite{Weber2011}, along with a connection from the internal provers to some very powerful external provers like E, SPASS, Vampire, CVC3 and Z3 through the Sledgehammer tool \cite{PaulsonLawrenceC.Blanchette2010}.
 
The \textit{Transfer} package was first released for Isabelle 2013-1 as a general mechanism for defining quotient types and transferring knowledge from the old `representation' type into the new `abstract' type \cite{huffman2013lifting}. However, their generalisation is not restricted to the definition of new quotient types, but allows the user to relate any two types by theorems of a specific shape called \textit{transfer rules}. Some of these rules can be defined automatically when the user defines a new quotient type, but the user is free to add them manually, provided that they prove a preservation theorem. Central to this package, the \textit{transfer} and \textit{transfer}$^{\prime}$ tactics try to automatically match the goal sentence to a new one related by either equivalence or implication, inferring this relation from the transfer rules.

We have taken full advantage of the generality of the transfer package as a means of automating the translation between sentences across domains which are related by what we consider an appropriate and general notion of \textit{structural transformation}. In Section \ref{transthy} we give an overview of our notion of transformation and how the tactics of the transfer package are useful mechanisms for exploiting the knowledge of a structural transformation.

\vspace*{-.15cm}
\section{Overall vision}\label{overall}
\vspace*{-.15cm}
The worlds of mathematical entities are interconnected. Numbers can be represented as sets, pairs of sets, lists of digits, bags of primes, etc. Some representations are only \textit{foundational} and the reasoner often finds it more useful to discard the representation for practical use (e.g., natural number 3 is represented by $\{ \emptyset, \{\emptyset\}, \{\emptyset, \{\emptyset\}\} \}$ in the typical ZF foundations, but this representation is rarely used in practice), and some are \textit{emergent}; they only come about after a fair amount of accumulated knowledge about the objects themselves, but are more helpful as reasoning tools (e.g., natural numbers as bags of primes). Overall, we think that there is no obvious notion of `better representation', and it's up to the reasoner to choose, depending on the task at hand. Thus, we envision a system where the representation of entities can be fluidly transformed.

We have looked at problems in discrete mathematics and the transformations commonly used for solving them. Below, we give one motivating example and show how we have mechanised the transformation in question inside Isabelle/HOL. Other motivating examples are briefly mentioned.

\subsection*{Numbers as bags of primes}\label{firstexample}
Let us start with an example of the role of representation in number theory. Consider the following problem:

\begin{problem}\label{prob1}
Let $n$ be a positive integer. Assume that, for every prime $p$, if $p$ divides $n$ then $p^2$ also divides $n$. Prove that $n$ is the product of a square and a cube.
\end{problem}
A standard solution to this problem is to take a set of primes $p_i$ such that $n = p_1^{a_1} p_2^{a_2} \cdots p_k^{a_k}$. Then we notice that the condition ``if $p$ divides $n$ then $p^2$ also divides $n$" means that $a_i \neq 1$, for each $a_i$. Then, we need to find $x_1, x_2, \ldots , x_k$ and $y_1, y_2, \ldots, y_k$ where \[(p_1^{x_1} p_2^{x_2} \cdots p_n^{x_k})^2  (p_1^{y_1} p_2^{y_2} \cdots p_n^{y_k})^3 = p_1^{a_1} p_2^{a_2} \cdots p_n^{a_k}\]
or simply \[2(x_1, x_2, \ldots , x_k) + 3(y_1, y_2, \ldots, y_k) = (a_1, a_2, \ldots, a_k).\]
Thus, we only need to prove that for every $a_i \neq 1$ there is a pair $x_i$, $y_i$ such that $2x_i + 3y_i = a_i$. The proof of this is routine.

The kind of reasoning used for this problem is considered standard by mathematicians. However, it is not so simple in current systems for automated theorem proving. The non-standard step is the `translation' from an expression containing various applications of the exponential function into a simpler form in a linear arithmetic of lists, validated by the fundamental theorem of arithmetic. 

The informal nature of the argument, in the usual mathematical presentation, leaves it open whether the reasoning is best thought as happening in an arithmetic of lists where the elements are the exponents of the primes, or perhaps a theory of bags (multisets) where the elements are prime numbers. The reader might find it very easy to fluidly understand how these representations match with each other and how they are really just different aspects of the same thing. Such ease supports our overall argument and vision: that to automate mathematical reasoning, we require a framework in which data structures are linked robustly by logically valid translations, where the translation from one to another is easily conjured up.

The \textit{numbers-as-bags-of-primes} transformation that links each positive integer to the bag of its prime factors is valid because there are operations on each side (numbers and multisets) that correspond to one another. For example, `divides' corresponds to `sub(multi)set', `least common multiple' corresponds to `union', `product' corresponds to `multiset addition', etc. Furthermore, all the predicates used in the statement of problem \ref{prob1} have correspondences with well-known predicates regarding bags of primes. Thus, the problem can be translated as a whole. Other representations may not be very productive, e.g., try thinking about exponentiation in terms of lists of digits.

Table \ref{tab1} shows more examples of number theory problems with their corresponding problem about multisets.
\vspace*{-.5cm}
\begin{table}
\setlength{\abovecaptionskip}{10pt plus 2pt minus 2pt}
\renewcommand{\arraystretch}{1.5}
\renewcommand{\tabcolsep}{6pt}
\begin{tabular}{|p{0.47\linewidth}|p{0.47\linewidth}|}\hline

{\bf Problem in $\N$} &
{\bf Problem in multisets} \\ \hline

{Prove that there is a unique set $\{x,y,z\}$ with different $x$, $y$, $z$ greater than 1, such that $xyz = 100$.} & 
{Prove that there is a unique way to partition $\{2,2,5,5\}$ into three different non-empty parts.} \\ \hline 


{Prove that in a set of 9 natural numbers, where none is divided by a prime larger than 6, there is a pair whose product is a perfect square.} &  
{Take 9 multisets whose only elements are $2$, $3$ and $5$. Prove that two of the multisets have multiplicities with the same parity.} \\ \hline 

\end{tabular}
\caption{Number theory problems and their multiset counterparts.} \label{tab1}
\end{table}

\vspace*{-1cm}
\subsection*{Numbers as sets}
 Many numerical problems have \textit{combinatorial proofs}. Theses are proofs where numbers are interpreted to be cardinalities of sets, and the whole problem can be converted to a problem about sets.

\textit{Enumerative combinatorics} studies how sets relate to their cardinalities. As such, its theorems provide the link that allows us to translate numerical problems into finite set-theoretical problems.

Table \ref{tab2} shows examples of arithmetic problems with their corresponding finite set theory problems. While the proofs of the numerical versions are not obvious at all (some of which are important results in basic combinatorics), the proofs of their finite set versions can be considered routine.
\vspace*{-.5cm}
\begin{table}[ht]
\setlength{\abovecaptionskip}{10pt plus 2pt minus 2pt}
\renewcommand{\arraystretch}{1.2}
\renewcommand{\tabcolsep}{6pt}
\begin{tabular}{|p{0.38\linewidth}|p{.56\linewidth}|}
\hline
{\bf Problem in $\N$} &
{\bf Problem in sets} \\ \hline
\centering
\multirow{3}{*}{$\displaystyle{n+1 \choose k+1} = {n \choose k} + {n \choose k+1}$} & \multirow{3}{\linewidth}{The set $\{x \subseteq \{0,1,\ldots,n\} : |x| = k+1\}$ can be partitioned into $2$ parts: those that contain element $n$ and those that don't.}\\ & \\ & \\
\hline
 \centering
\multirow{3}{*}{$\frac{n(n+1)}{2} = \displaystyle 1+ 2+ \cdots + n$} & \multirow{3}{\linewidth}{The set $\{x \subseteq \{0,1,\ldots,n\} : |x| = 2\}$ can be partitioned into $n$ parts $X_1,X_2,\ldots,X_n$ where the largest element of each $x \in X_i$ is $i$.}\\ & \\ & \\
\hline \centering
\multirow{4}{*}{$2^{n+1} - 1 = \displaystyle\sum_{i=0}^n 2^i$} & \multirow{4}{\linewidth}{The power set of $\{0,1,\ldots,n\}$, excluding the empty set, can be partitioned into $n$ parts $X_1,X_2,\ldots,X_n$ where the largest element of each $x \in X_i$ is $i$.}\\ & \\ & \\ & \\
\hline \centering
\multirow{3}{*}{$2^n = \displaystyle\sum_{i=0}^n  {n \choose i}$} & \multirow{3}{\linewidth}{The power set of $\{1,\ldots,n\}$ can be partitioned into $n+1$ parts $X_0,X_1,\ldots,X_{n}$ where $|x| = i$ for every $x \in X_i$.}\\ & \\ & \\
\hline
\end{tabular}
\caption{Numerical problems and their set counterparts.} \label{tab2}
\end{table}

\vspace*{-1cm}
\subsection*{Interconnectedness}
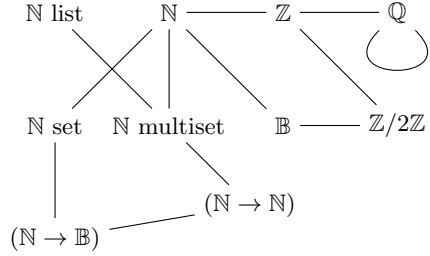
\begin{figure}[htb]\small\centering
\begin{tikzpicture}[node distance=1.5cm, auto]
  \node (list) {$\mathbb{N}$ list};
  \node (nat) [right of=list] {$\mathbb{N}$};
  \node (mset) [below of=nat] {$\mathbb{N}$ multiset};
  \node (set) [below of=list] {$\mathbb{N}$ set};
  \node (int) [right of=nat] {$\mathbb{Z}$};
  \node (rat) [right of=int] {$\mathbb{Q}$};
  \node (int2) [below of=rat] {$\mathbb{Z}/2\mathbb{Z}$};
  \node (bool) [below of=int] {$\B$};
  \node (fset) [below of=set] {$(\mathbb{N} \to \B)$};
  \node (fmset) [below right of=mset] {$(\mathbb{N} \to \mathbb{N})$};
  \draw[-] (mset) to node {} (nat);
  \draw[-] (nat) to node {} (set);
  \draw[-] (list) to node {} (mset);
 
  \draw[-] (mset) to node {} (fmset);
  \draw[-] (set) to node {} (fset);
  \draw[-] (fset) to node {} (fmset);
  
  \draw[-] (nat) to node {} (int);
  \draw[-] (nat) to node {} (bool);
  \draw[-] (int) to node {} (rat);
  \draw[-] (int) to node {} (int2);
  \draw[-] (bool) to node {} (int2);
  \draw[-, loop below, out = 315, in = 225, distance=1cm] (rat) to node {} (rat);
\end{tikzpicture}
\caption{\footnotesize{Nodes stand for theories connected by transformations useful in discrete mathematics. Apart from the aforementioned transformations, it includes other simpler ones. Actually, some of these transformations, such as that connecting $\N$ list and $\N$ set, are polymorphic, but presented in the diagram as relating only to type $\N$. This is because the numbers-as-bags-of-primes transformation is not polymorphic.}} \label{figure} 
\end{figure}

We want to stress the importance of having fluidity of representations. For example, we talked about the ease with which we could think that the \textit{numbers-as-bags-of-primes} transformation is actually a transformation of numbers to a theory of lists, where elements of the list are the exponents of the ordered prime factors. Inspired by this, we have mechanised many other simple transformations, but whose composition allows us to translate fluently from one representation to another. Our global vision of transformations useful in discrete mathematics, which we have mechanised\footnote{These can be found in http://homepages.inf.ed.ac.uk/s1052074/AutoTransfer/. They are updated regularly.}, is represented in Figure \ref{figure}. It is worth mentioning that the diagram is not commutative and that it abstracts logical relations (information may be lost, so some paths can only be traversed in one direction).

In the next section we show how a notion of transformation that accounts for this kind of correspondence between structures can be applied in formal proofs using Isabelle's Transfer tool.

\vspace*{-.15cm}
\section{On Transformations and the Transfer tool}\label{transthy}
\vspace*{-.2cm}
In this section we give a brief overview of a very general theory of transformations. We do not claim originality of the essence of this theory. However, we believe that the presentation we give brings clarity to the problem. We explain how Isabelle's Transfer tool relates to it. Consider the following definitions:
\begin{definition} A \textbf{domain} is a class of entities and a set of types, where each entity of the domain corresponds to exactly one type. \end{definition}
\begin{definition} A \textbf{transformation} from a domain $\mathscr{X}$ to a domain $\mathscr{Y}$ is a collection $\mathscr{R}$ where every $R \in \mathscr{R}$ is a relation $R : X \to Y \to \B$ between a type $X$ of domain $\mathscr{X}$ and a type $Y$ of domain $\mathscr{Y}$.\footnote{$\B$ stands for type of booleans.}\end{definition}
This relational notion of a transformation makes it possible to account for partial and multivalued mappings in a logic like Isabelle's HOL. 

We consider a \textit{structure} to be the class conformed by all the entities of the closure of a domain under a set of type constructors. In this work, we focus on structures containing type $\B$, generated with the \textit{function type constructor} $\to$, because the basis of a higher order logic can be fully expressed under such a structure. Then, if our domain has entities of types $A$ and $B$, its structure under $\to$ has all the entities of types $A \to B$, $A \to B \to A$, etc.

Preservation of structure is captured with the use of structural \textit{relators}, which can be thought of as rules for extending relations (transformations) to the structures of their domains. In particular, given that our work is based on Isabelle/HOL and on the Transfer package, we focus on one relator.

\begin{definition} The \textbf{standard functional extension} of two relations $R_A : A \to A' \to \B$ and $R_B : B \to B' \to \B$ (written $R_A \Mapsto R_B$) is a relation that relates two functions $f:A \to B$ and $f':A' \to B'$ whenever they satisfy the following property: \[\forall\, a : A.\;\, \forall\, a' : A'.\;\, \left[R_A \, a \, a' \imp R_B \, (f \, a) \, (f' \, a') \right]\]
\end{definition} 
We call the operator $\Mapsto$ the \textbf{standard function relator}. Intuitively, $(R_A \Mapsto R_B)\, f\, g$ means that $f$ and $g$ send arguments related by $R_A$ to values related by $R_B$. This relator allows us to express how functions (and by extension relations) map to each other in a way that the structure of the domain is preserved. 

For the numbers-as-bags-of-primes transformation, consider relation $\mathcal{F} : \N \to \N\; \texttt{multiset} \to \B$, that relates every positive integer with the multiset of its prime factors.

\begin{example}  Let $\ast : \N \to \N \to \N$ be the usual multiplication and \linebreak $\uplus : \N \; \texttt{multiset} \to \N \; \texttt{multiset} \to \N \; \texttt{multiset}$ the `addition' of multisets (in which the multiplicities are added per element). Then we have $(\mathcal{F} \Mapsto (\mathcal{F} \Mapsto \mathcal{F}))\, \ast \, \uplus$ (also written $(\mathcal{F} \Mapsto \mathcal{F} \Mapsto \mathcal{F})\, \ast \, \uplus$). 

\noindent Note that, by expanding the definition of\, $\Mapsto$ in $(\mathcal{F} \Mapsto (\mathcal{F} \Mapsto \mathcal{F}))\, \ast \, \uplus$ we get \[\forall\,n_1.\, \forall\,N_1.\; \mathcal{F}\, n_1\, N_1 \imp \left(\forall\,n_2.\, \forall\, N_2.\; \mathcal{F}\, n_2\, N_2 \imp \mathcal{F}\, (n_1 \ast n_2)\, (N_1 \uplus N_2)\right)\]
which is equivalent to \[\forall\,n_1, n_2.\; \forall\,N_1, N_2.\; \left(\mathcal{F}\, n_1\, N_1 \wedge \mathcal{F}\, n_2\, N_2\right) \imp \mathcal{F}\, (n_1 \ast n_2)\, (N_1 \uplus N_2)\] This demonstrates how nesting the operator\, $\Mapsto$ preserves its intuitive definition: `related arguments map to related values'. In this particular case, this is true due to the law of exponents $p^a p^b = p^{a+b}$.
\end{example}

Furthermore, the matching of relations can also be expressed with the help of $\Mapsto$, using a boolean relation, as demonstrated by the example below with equivalence (boolean equality) $\texttt{eq} : \B \to \B \to \B$.
\begin{example} Let $\texttt{div}: \N \to \N \to \B$ be the relation such that $\texttt{div} \; n \, m$ whenever $n$ divides $m$ (also written $n|m$), and $\subseteq: \N \; \texttt{multiset} \to \N \; \texttt{multiset} \to \B$ the relation such that $a \subseteq b$ whenever the multiplicity of each element of $a$ is lesser or equal to its multiplicity in $b$. Then, we have $(\mathcal{F} \Mapsto \mathcal{F} \Mapsto \texttt{eq}) \;\, \texttt{div} \, \subseteq$, because $n$ divides $m$ if and only if every prime is contained at least as many times in the multiset-factorisation of $m$ as it is in $n$.
\end{example}

Logical matches (preservation of truth values) can also be expressed across structures, e.g.,\, $(\texttt{eq} \Mapsto \texttt{eq} \Mapsto \texttt{eq}) \; \texttt{imp} \; \texttt{imp}$\, represents that implication $\texttt{imp}$ preserves truth if its arguments are replaced by equivalent ones. Other interesting logical matches can be expressed as well. 

The general notion of transformation above tells us how theories will relate to one another. Isabelle's Transfer method is an algorithm for transforming a sentence using knowledge about one of these transformations. The simple standard function relator is at the basis of the method. We give a short introduction next.

\subsection{Transforming sentences with the Transfer tool}\label{transformingproblems}
\vspace*{-.1cm}
When trying to prove a sentence $\beta$ we want to find another sentence $\alpha$ such that $\alpha \imp \beta$, along with a proof for $\alpha$. In particular, if $\beta$ talks about a domain $B$ and we know a structural transformation from a domain $A$ to $B$, we might be able to find an $\alpha$ about $A$, such that $\alpha \imp \beta$.

Isabelle's \textit{Transfer} tool provides a method for finding such $\alpha$. The user has to provide theorems of the forms $R_1\, a\, b$ or $(R_1 \Mapsto R_2)\, f\, g$ (and their proofs), i.e., instances of a structural transformation, and the tactics $\texttt{transfer}$ and $\texttt{transfer}'$ will try to automatically infer a sentence $\alpha$ such that $\alpha \bicond \beta$ (in the case of $\texttt{transfer}$), or a weaker one such that $\alpha \imp \beta$ (in the case of $\texttt{transfer}'$). 

Recall that the intuitive interpretation of $(R_1 \Mapsto R_2)\, f \, g$ is `arguments related by $R_1$ are mapped to values related by $R_2$ by $f$ and $g$. Thus, the first step of the transfer method is to search for a theorem of the structural transformation with the shape $(R_1 \Mapsto \texttt{eq})\, p \, q$ in the case of $\texttt{transfer}$ and $(R_1 \Mapsto \texttt{imp})\, p \, q$ in the case of $\texttt{transfer}'$, where $q$ is the property wrapping the sentence we want to prove. Finding it would imply that we can replace $q$ by $p$ provided that we can find that their arguments are related by $R_1$. Thus, the method searches recursively for rules in the structural transformation to prove this. The algorithm is analogous to type inference. It is based on the following derivation rules:

\begin{prooftree}
\AxiomC{$\mathscr{A}^{\ast}_{\mathscr{C}} \vdash (R_1 \Mapsto R_2) \, f\, g$}
\AxiomC{$\mathscr{A}^{\ast}_{\mathscr{C}} \vdash R_1\, x \, y$}\RightLabel{elim}
\BinaryInfC{$\mathscr{A}^{\ast}_{\mathscr{C}} \vdash R_2\, (f\, x) \, (g\, y)$}
\end{prooftree}
\begin{prooftree}
\AxiomC{$\mathscr{A}^{\ast}_{\mathscr{C}}, \def\extraVskip{2pt} R_1 \, x \, y \vdash R_2\, (f \, x) \, (g \, y)$}\RightLabel{intro}
\UnaryInfC{$\mathscr{A}^{\ast}_{\mathscr{C}} \vdash (R_1 \Mapsto R_2) \; (\lambda x.\, f\, x)\; (\lambda y.\, g\, y)$}
\end{prooftree}

where $\mathscr{A}^{\ast}_{\mathscr{C}}$ represents knowledge about the structural transformation. Practically, the user provides knowledge specific to this transformation (a set of theorems called \textit{transfer rules}), and the algorithm includes in the search other general transfer rules such as $(\texttt{eq} \Mapsto \texttt{eq} \Mapsto \texttt{eq})\; \texttt{imp} \; \texttt{imp}$. For more details of the actual implementation of the algorithm see \cite{huffman2013lifting}.

\vspace*{-.15cm}
\section{Mechanising transformations in Isabelle's HOL}
\vspace*{-.15cm}
In Section \ref{firstexample} we presented some problems in discrete mathematics which involve structural transformation. We have mechanised the transformations by proving the necessary transfer rules. The transfer tool allows us to use the transformations in proofs. 

In this section we present a couple of examples from a larger catalogue of the transformations we have mechanised in Isabelle. The transformations we have formalised, as suggested in Figure \ref{figure}, are the following: 

\begin{enumerate}
\item \textbf{numbers-as-bags-of-primes}, where each natural number is related to the multiset of its prime factorisation. 

\item \textbf{numbers-as-sets}, where numbers are related to sets by the cardinality function. 

\item \textbf{sets-as-$\B$-functions}, where sets are seen as boolean-valued functions. 

\item \textbf{multisets-as-$\N$-functions}, where multisets are seen as natural-valued functions\footnote{This one is actually by construction using \texttt{typedef} and the Lifting package, which automatically declares transfer rules from definitions lifted by the user from an old type to the newly declared type.}. 

\item \textbf{sets-as-lists}, where sets are related to lists of their elements. 

\item \textbf{bits-from-integers}, where type \texttt{bit} is created as an abstract type from the integers. 

\item \textbf{bits-as-booleans}, where bits are matched to booleans. 

\item \textbf{$\mathbb{Q}$-automorphisms}, where rational numbers are stretched and contracted, parametric on a factor. 

\item \textbf{zero-or-some}, where natural 0 is related to bit 0 and positive natural numbers are related to bit 1. 

\item \textbf{multisets-as-lists}, where multisets are related to lists of their elements. 

\item \textbf{set-to-multiset}, where the functional representations of multisets and sets are related (this one, we get it for free from the zero-or-some transformation). 

\item \textbf{naturals-as-integers}, where naturals are matched to integers (this one was built by the developers of the transfer package, not us). 

\item \textbf{integers-as-rationals}, where integers are matched to rational numbers. Notice that composition of transformations leads to other natural transformations, like the simple relation between sets and multisets.\footnote{The mechanisation of these transformations have been submitted to the Archive of Formal Proofs, along with some examples of their use.} 
\end{enumerate}

Every transformation starts with a declaration and proof of \textit{transfer rules}, which are sentences satisfied by the structural transformation.

\subsection{Numbers as bags of primes}
\vspace*{-.1cm}
The relation at the centre of this transformation is $\mathcal{F} : \N \to \N\;\texttt{multiset} \to \B$, which relates every positive number to the multiset of its prime factors. It is defined as follows: $\mathcal{F}\, n\, M$ holds if and only if \[(\forall\, x.\; \texttt{count}\, M \, x > 0 \imp \texttt{prime} \, x) \land n = \prod_{x \in M} x^{\texttt{count}\; M \, x}\] 

The most basic transfer rules (instances of the structural transformation) are theorems such as $\mathcal{F}\; 6\; \{2,3\}$, whose proof are trivial calculations. Moreover, from the Unique Prime Factorisation theorem we know that $\mathcal{F}$ is bi-unique. Thus, we know that 
\[(\mathcal{F} \Mapsto \mathcal{F} \Mapsto \texttt{eq})\, \texttt{eq}\, \texttt{eq}\] i.e., that equality is preserved by the transformation. 

From the fact that every positive number has a factorisation we have
 \begin{align*}
&((\mathcal{F} \Mapsto \texttt{revimp}) \Mapsto \texttt{revimp})\; \forall_{>0} \; \forall \hspace*{1cm} 
&((\mathcal{F} \Mapsto \texttt{revimp}) \Mapsto \texttt{revimp})\; \exists \; \exists_p \\
&((\mathcal{F} \Mapsto \texttt{eq}) \Mapsto \texttt{eq})\; \forall_{>0} \; \forall_p \hspace*{1cm} 
&((\mathcal{F} \Mapsto \texttt{eq}) \Mapsto \texttt{eq})\; \exists_{>0} \; \exists_p 
\end{align*}
where $\texttt{revimp}$ is reverse implication, $\forall_p$ is the bounded quantifier representing `for every multiset where all its elements are primes' and $\forall_{>0}$ is the bounded quantifier representing `for every positive number', and similarly for $\exists_p$ and $\exists_{>0}$. The mechanised proofs of these sentences follow relatively straightforward from the Unique Prime Factorisation theorem which is already part of Isabelle's library of number theory.

Furthermore, we proved the following correspondences of structure: 
\begin{align*}
&(\mathcal{F} \Mapsto \mathcal{F} \Mapsto \mathcal{F}) \, \ast\; \uplus
\hspace*{2cm}
&(\mathcal{F} \Mapsto \mathcal{F} \Mapsto \mathcal{F}) \, \texttt{lcm}\; \cup \\
&(\mathcal{F} \Mapsto \mathcal{F} \Mapsto \mathcal{F}) \, \texttt{gcd}\; \cap
\hspace*{2cm}
&(\mathcal{F} \Mapsto \mathcal{F} \Mapsto \texttt{eq}) \, \texttt{div} \subseteq 
\\
&(\mathcal{F} \Mapsto \texttt{eq} \Mapsto \mathcal{F})\; \texttt{exp} \; \texttt{smult}
\hspace*{2cm}
&(\mathcal{F} \Mapsto \texttt{eq})\; \texttt{prime} \; \texttt{sing}
\end{align*}

\subsubsection*{Application in proofs}\mbox{}
\vspace*{.2cm}
\newline
We formalised the proof of problem \ref{prob1}:
\begin{center}
\emph{
Let $n$ be a positive integer. Assume that, for every prime $p$, if $p$ divides $n$ \break then $p^2$ also divides it. Prove that $n$ is the product of a square and a cube.
}
\end{center}

\noindent Formally, we state this as 
\begin{align*}\forall\, n>0.\; (\forall\, p>0.\; \texttt{prime}\; p \,\land\, p\; \texttt{div}\; n \imp p^2 \;\texttt{div}\; n&) \\ \imp (\exists\, a>0.\; \exists\, b>0.\; a^2 \ast b^3 = n&)
\end{align*}
Notice that the quantifiers of $n$ $p$, $a$ and $b$ are bounded (greater than 0). This is not necessary (e.g., $p$ is prime, so it is redundant to say that it is positive), but it is convenient for the proof. If we want a proof for the unbounded version (which is also a theorem) we can divide in cases, when $n=0$ and when $n>0$. The case for $n = 0$ is trivial because then $a= 0$ and $b = 0$ are solutions. Thus, we prove prove directly the case for $n > 0$.

When we apply the transfer method to the sentence we get the following sentence about multisets:
\[
\forall_p\, n.\;
     (\forall_p \, p.\; \texttt{sing}\; p \land p \subseteq n \imp 2 \cdot p \subseteq n) \imp
         (\exists_p\, a.\; \exists_p\, b.\; 2 \cdot a + 3 \cdot b = n)
\]
where $\forall_p$ is the universal quantifier bounded to prime numbers, and operator $\cdot$ represents the symmetric version of the multiplication previously referred to as $\texttt{smult}$ (we present it as we do for reading ease).

The premise $(\forall_p \, p.\; \texttt{sing}\; p \land p \subseteq n \imp 2 \cdot p \subseteq n)$ is easily proved to be equivalent to $\forall\, q. \; \texttt{count}\, n\, q \neq 1$. Then it is sufficient to show 
\[
\forall_p\, n.\;
     (\forall\, q. \; \texttt{count}\, n\, q \neq 1) \imp
         (\exists_p\, a.\; \exists_p\, b.\; 2 \cdot a + 3 \cdot b = n)
\]
With a bit of human interaction, this can further be reduced to proving that, for every element of $n$, its multiplicity $n_i$ (which the premise says is different from 1) can be written as $2 a_i + 3 b_i$, or formally: \[ \forall\,n_i :\N.\; n_i \neq 1 \imp \exists\,a_i.\;\exists\,b_i.\; 2 a_i + 3 b_i = n_i\] This problem can actually be solved in a decidable part of number theory (Presburger arithmetic), for which there is a method implemented in Isabelle.

\subsection{Numbers as sets}
\vspace*{-.1cm}
At the centre of this transformation is the relation $\mathcal{C}$ where $\mathcal{C}\, A\, n $ holds if and only if\, $\texttt{finite} \,A \wedge \texttt{card}\, A = n$.

We first prove trivial cardinality properties like\, $\mathcal{C} \, \{1 \cdots n\}\, n$, which allows us to consider standard representatives of numbers.

This relation is right-total but not left total, so we have the following two rules:
\begin{align*}
((\mathcal{C} \Mapsto \texttt{imp}) \Mapsto \texttt{imp})\; \forall \; \forall
\hspace*{1.5cm}
((\mathcal{C} \Mapsto \texttt{eq}) \Mapsto \texttt{eq})\; \forall_{\texttt{fin}} \; \forall
\end{align*}
where $\forall_{\texttt{fin}}$ is the universal quantifier restricted to finite sets.
Furthermore, the relation is left-unique but not right-unique, so we have
\begin{align*}
(\mathcal{C} \Mapsto \mathcal{C}  \Mapsto \texttt{imp})\; \texttt{eq} \; \texttt{eq}
\hspace*{1.5cm}
(\mathcal{C} \Mapsto \mathcal{C}  \Mapsto \texttt{eq})\; \texttt{eqp} \; \texttt{eq}
\end{align*}
where $\texttt{eqp}$ is the relation of being equipotent, or bijectable.

Then, we have the following rules for the structural correspondence:
\begin{align*}
(\mathcal{C} \Mapsto \mathcal{C})&\; \texttt{Pow}\; (\lambda x.\, 2^x)
\\
(\mathcal{C} \Mapsto \texttt{eq} \Mapsto \mathcal{C})&\; \texttt{n-Pow}\; \left(\lambda\, n\, m.\, \textstyle{n \choose m}\right)
\\
(\mathcal{C} \Mapsto \mathcal{C} \Mapsto \texttt{imp})&\, \subseteq\; \leq
\\
(\mathcal{C} \Mapsto \mathcal{C} \Mapsto \mathcal{C} \Mapsto \texttt{imp})&\; \texttt{disjU} \; \texttt{plus}
\end{align*}
where $\texttt{n-Pow}\, S\, n$ is the operator that takes the set of subsets of $S$ that have cardinality $n$. Also, $\texttt{disjU}\; a\; b\; c$ means $\texttt{disjoint} \; a\; b \wedge a \cup b = c$ and $\texttt{plus}$ is the predicative form of operator $+$.

We have mechanised combinatorial proofs, like the ones for the problems given in Table \ref{tab2}, of theorems using this transformation.

\vspace*{-.15cm}
\section{Automated change of representation}
\vspace*{-.15cm}
We have built a tactic that searches within the space of representations given a set of transformations. Then it tries to reason about each representation. Our goal is for it to embody our vision presented in Section \ref{overall}. This is work in progress, but we address some simple requirements that we have already implemented and present our observations.

\vspace*{-.15cm}
\subsection{Transformations as sets of transfer rules}
\vspace*{-.1cm}
As described in Section \ref{transthy}, we consider a transformation as a set of `base' relations, and a structural extension of them. Then, \textit{knowing} a transformation means knowing instances where the relations and their extensions (with respect to relators such as $\Mapsto$) hold. These instances of knowledge are what the Transfer package calls \textit{transfer rules}. They are theorems that the user has to prove and, with enough of them provided, the transfer method will transform the goal to an equivalent, or stronger sentence in another domain.

In the traditional use of the Transfer method, there is a single attribute that encompasses all transfer rules. Given a goal, the Transfer method will try to derive an equivalent or stronger subgoal using all the rules with such attribute, with a simple inference mechanism (described in briefly in Section \ref{transformingproblems} and more detailed in \cite{huffman2013lifting}).  We have packaged each of the transformations described in Figure \ref{figure} as a set of transfer rules. Then, our tactic applies the transfer method one transformation at a time.

\vspace*{-.15cm}
\subsubsection*{Transformation-specific language.}
Each transformation has a set of definitions that are linked by the transfer package. Some of them are defined only for use of the transformation, like \texttt{disjU} and \texttt{plus} (the predicative version of disjoint union of sets and addition of natural numbers, respectively), or bounded quantifiers. These are necessary for the transfer method to find matches, but theorems will not generally be stated in such terms. Our tactic normalises the language of the goal to suit the specific transformation that is going to be applied.

\vspace*{-.15cm}
\subsection{Reversing transformations}
\vspace*{-.1cm}
We have implemented a tool to automatically \textit{reverse} transformations. Let us explain this.

If we want to transform a sentence $p\, a$ to an equivalent one, the Transfer method will search for transfer rules $(R \Mapsto \texttt{eq})\, q \, p$ and  $R\, b\, a$ for some $R$, $q$ and $b$. If found, it will transform the sentence to the equivalent one $q\, b$. The fact that the sentences are equivalent means that if we had started with $q \, b$ as a goal, it would have been valid to transform it to $p\, a$. This means that, in theory, the same transfer rules can be used to do inference in one direction or the other, at least when the rules are regarding equivalence. The Transfer method does not do so: if one wants to use a transformation in both directions one has to define two distinct transformations, i.e., two distinct sets of transfer rules (in our example above one needs transfer rules $(R^{\prime} \Mapsto \texttt{eq})\, p \, q$ and  $R^{\prime}\, a\, b$, where $R^{\prime}$ is the reverse of $R$). A transfer rule always has a `reverse' version (although only equivalent ones retain full information), so we should be able to get these automatically. We have built a conversion tool that, given a set of transfer rules, will generate all their reverse rules (in a logically valid way, i.e., the reverse version is always equivalent to the original).

Our program uses the following rewrite rules:
\begin{align*}
R\, a\, b &\Rightarrow (\texttt{swap}\, R) \, b\, a \\ 
\texttt{swap}\, (R_1 \Mapsto R_2) &\Rightarrow (\texttt{swap}\, R_1 \Mapsto \texttt{swap}\, R_2)
\end{align*}
where $\texttt{swap}$ simply swaps the place of the arguments of a function. It is easy to see that these rules are valid. Moreover, $\texttt{swap}\, R$ equals $R$ when $R$ is symmetric, which means that in some relations we can drop the $\texttt{swap}$ function. Thus, our program drops \texttt{swap} from $\texttt{eq}$ and turns $\texttt{swap}\; \texttt{imp}$ and $\texttt{swap}\; \texttt{revimp}$ into $\texttt{revimp}$ and $\texttt{imp}$, respectively.

By reversing every transformation we can traverse every path in Figure \ref{figure} in any direction (which does not mean that every sentence has a transformation to an equivalent one).

\vspace*{-.15cm}
\subsection{Search between representations}
\vspace*{-.1cm}
Our tactic searches the space of representations by applying each transformation, then reasoning within the theory where it arrived, and, if there are still open subgoals it will repeat the process iteratively. 

Recall that transformations are relational. As such, the process is non-\linebreak deterministic for each transformation, so there will be many branches per transformation. Apart from being non-deterministic, the transfer method will allow transformations of a sentence where some matches are left open, i.e., in the place of some constant we get a \textit{schematic variable} that the user can instantiate manually, and prove their validity with the new instantiation. This can be handy, but our tactic favours branches with the lowest number of open subgoals, thus favouring complete matches; e.g., matches that will not leave any proof obligations open. 

We have also noticed that the order in which the transformations are searched is crucial and have set an ad hoc order that favours the transformations we consider more interesting. Heuristics deserve further work, but that remains as a task for the future.

\vspace*{-.15cm}
\subsubsection*{Discarding false representations.}
\vspace*{-.1cm}
Recall that our transformations do not necessarily yield equivalent sentences when applying the transfer algorithm (unless we restrict it to do so). Actually, the \textit{numbers-as-sets} transformation can only be applied in useful ways if we allow the reduction of the goal to a strictly stronger subgoal (because, e.g., $A \subseteq B$ implies that $|A| \leq |B|$, but not the  other way around, meaning that we can prove $|A| \leq |B|$ by showing first $A \subseteq B$, but we cannot prove $A \subseteq B$ by showing $|A| \leq |B|$). This can lead to false subgoals. Thus, our tactic calls the counterexample checker nitpick \cite{blanchette2010nitpick} and discards branches where a counterexample is found for one of its goals.

\subsection{Overview}
\vspace*{-.1cm}
In a single step in the search, our tactic does the following:
\begin{enumerate}
\item Normalise to transformation-specific language.
\item Apply transformation.
\item If working with a transformation that generates a stronger subgoal, search for counterexamples and discard if they are found.
\item Apply \texttt{auto} tactic to transformed sentence.
\end{enumerate}

The tactic can be applied recursively to search for a transformation to a domain more than one step away. When searching, the obvious stop condition is that the theorem has been proved, although there can be other good reasons to stop in a domain to allow the user to reason interactively.

Each of the 4 steps mentioned can have plenty of branches, so there is search involved. Branches with the least number subgoals are favoured, and the order in which the transformations are applied matters, but there are no clever heuristics involved. 

Even though our observations about the trace of the search have led us to the current design and implementation of the tactic, the design is not yet complete and its implementation (although functional) is very much subject to change. There are still open questions regarding what search strategies, \textit{stop} conditions, and reasoning tactics (between transformations) are the best, because these are subject to what evaluation criterion we should use. In Section \ref{concl} we discuss why this is problematic and how we are confronting it.

\vspace*{-.15cm}
\section{Related Work}
\vspace*{-.2cm}
Although representation is widely recognised as a crucial aspect of reasoning, to our knowledge there has been no attempt to incorporate the \textit{automatic} search of representation into reasoning tools.
\vspace*{-.2cm}
\subsection*{Institutions and HETS}
\vspace*{-.1cm}
The concept of Institution was introduced to as a general notion of logical system \cite{goguen1992institutions}. The Heterogeneous Tool Set (HETS) \cite{hets07} was developed mainly to manage and integrate heterogeneous specifications. Based on the theory of Institutions, it links various logics, including Isabelle's HOL and FOL, and provides a way of translating between them. The uses of HETS have been to bring together various aspects of complex systems where different programming languages and reasoning tools are used for different parts of the system. We do not know of any uses of HETS where heterogeneity is taken advantage of as a means of finding proofs in one representation where other representations fail.
\vspace*{-.2cm}
\subsection*{Little Theories and IMPS}
\vspace*{-.1cm}
``Little Theories'' is the notion that reasoning is best done when it is modular \cite{farmer}. IMPS is a an interactive proof system implemented based on the principles of Little Theories \cite{farmerimp}. The modules, or `little theories' of IMPS are small axiomatic theories connected by \textit{theory interpretation}. Thus, it concerns different levels of abstraction of a theory, and not directly representation of the entities of the theory.
\vspace*{-.2cm}
\subsection*{Uses of the Transfer package}
\vspace*{-.1cm}
The use of the Transfer package has changed how new quotient types and subtypes are defined. This is what the \textit{Lifting} package does \cite{huffman2013lifting}. As part of the lifting package, there is a way of automatically transferring definitions from an old type to a new type (e.g., multisets are defined as an abstract type from the type of $\N$-valued functions).

The Lifting package has been the main application of the Transfer package, although the generality of their approach is acknowledged by the developers. Embodying this generality, they have built an Isabelle theory of transference from integers to natural numbers, very much in the spirit of the various transformations we have built ourselves.

\vspace*{-.15cm}
\section{Evaluation, Future Work and Conclusion}\label{concl}
\vspace*{-.2cm}
The main contributions presented in this paper are:
\begin{itemize}
\item We mechanised various useful transformations observed in proofs of discrete mathematics.
\item We have proved example theorems using these transformations.
\item We have identified some requirements for search over the space of representations, and implemented both a tool (for reversing transformations) and a tactic fulfilling the requirements.
\end{itemize}

Our tactic has yet to be evaluated properly. Below we examine some of the difficulties associated with this task.

What makes one proof better than another? There is no definite answer for this question. Simple measures, such as length, are important, but unsatisfactory as a whole. At the very least, we can agree that some proof is better than no proof. Thus, the simplest scenario for evaluation would be that in which our tactic that reasons within many representations finds proofs which cannot be found otherwise. Unfortunately, the current state of automatic theorem provers does not seem to be conducive to this. All the examples in which we have tested our techniques belong to either of the following classes: 
\begin{enumerate}
\item They are so simple that they can be proved automatically\footnote{using Isabelle tactics like \texttt{auto}} without the need of a transformation.
\item They are too complicated and require an intervention from the user to complete the proof, even after automatically applying a transformation.\footnote{The examples of this second (more interesting) class have been selected from either maths textbooks for undergraduate students, or from training material for contests such as the Mathematical Olympiads.}
\end{enumerate}
Thus, the proof-or-no-proof criterion is not applicable . Then, it is necessary to work on close analysis of interactive proofs with transformations and without them.

A venture for future research is the potential application of this framework for the transformation of geometric problems into algebraic representations, e.g., Gröbner bases \cite{buchberger1998grobner}, where there has been plenty of success in automated reasoning, or into SAT/SMT, which also have been an area of success in automation.\footnote{We thank the anonymous referees of this paper for suggested these possibilities. They remain as future work.}

Interestingly, we have an example (Pascal's theorem) that belongs to the class of problems where Isabelle's automatic tactics can find a proof, but where its proof using a transformation deserves attention. It is provable automatically (from the definition of the \texttt{choose} operator included in Isabelle's combinatorics library by its developers), but can be transformed using the numbers-as-sets transformations and proved only interactively there. Arguably, a combinatorial proof could be highly valued by mathematicians (or a scientist who analyses proofs), making this an example where the interactive proof deserves equal, or even more, attention than the automatic proof. 

Furthermore, even in the case in which we had automatic proofs using the usual tactics (like Pascal's theorem, mentioned above), we have to consider that these tactics depend on background knowledge (in our case, this amounts to Isabelle's libraries, which have been vastly populated by users). This raises the question: are there ways in which we can measure success independently of the background theories? We think that this is partially achievable by building simpler theories, with some equal level of measurable simplicity, and testing tactics that incorporate representational change there. Even if impractical by itself, this might bring some scientific insight that might lead to better reasoning tactics and theorem provers in the future.

\bibliographystyle{plain}
\end{document}